%% file: 42.tex
\documentclass[runningheads]{llncs}
\usepackage[T1]{fontenc}
\usepackage{graphicx}
\usepackage{pdfpages}
\usepackage{orcidlink}
\begin{document}
\title{Quechua Speech Datasets in Common Voice: \\
The Case of Puno Quechua}
%
%
\author{
Elwin Huaman\inst{1}\orcidlink{0000-0002-2410-4977} \and
Wendi Huaman\inst{2} \and
Jorge Luis Huaman\inst{2} \and
Ninfa Quispe\inst{2}
}
\authorrunning{E. Huaman et al.}
%
\institute{
University of Innsbruck, Austria \\
    \email{elwin.huaman@uibk.ac.at} \and 
National University of Altiplano Puno, Peru \\
    \email{wendi.huamanquispe@gmail.com},
    \email{jorgellhq@gmail.com}
}
\maketitle              
\begin{abstract}
Under-resourced languages, such as Quechuas, face data and resource scarcity, hindering their development in speech technology. To address this issue, Common Voice presents a crucial opportunity to foster an open and community-driven speech dataset creation.
This paper examines the integration of Quechua languages into Common Voice. We detail the current 17 Quechua languages, presenting Puno Quechua (ISO 639-3: qxp) as a focused case study that includes language onboarding and corpus collection of both reading and spontaneous speech data.
Our results demonstrate that Common Voice now hosts 191.1 hours of Quechua speech (86\% validated), with Puno Quechua contributing 12 hours (77\% validated), highlighting the Common Voice's potential. We further propose a research agenda addressing technical challenges, alongside ethical considerations for community engagement and indigenous data sovereignty. Our work contributes towards inclusive voice technology and digital empowerment of under-resourced language communities.
\end{abstract}
\input{chapter/1-intro}
\input{chapter/2-quechua-languages}

\input{chapter/3-case-study}
\input{chapter/4-results}
\input{chapter/5-conclusion}
\input{chapter/6-acknowledgment}
\bibliographystyle{splncs04}
\bibliography{42}

\end{document}

%% file: chapter/1-intro.tex
\section{Introduction}
\label{sec:intro}
The digital landscape for indigenous languages, especially those with limited written traditions, presents challenges and opportunities. A well-know challenge in language technology is the lack of data and linguistics resources in under-resourced languages~\cite{BesacierBKS14a,HuamanHH22}, such as the Quechua languages~\cite{HuamanHH23}. At the same time, initiatives such as Mozilla's Common Voice project offer an opportunity to create and expand speech datasets for these languages~\cite{ArdilaBDKMHMSTW20}.

Among the +40 Quechua language varieties identified by ISO 693-3\footnote{\url{https://iso639-3.sil.org/code/que}} and Ethnologue\footnote{\url{https://www.ethnologue.com/language/que/}} as Active, Stable, or Institutional, only four are taken into account by the Government of Peru~\cite{LOdP2018}. Even these are not treated as distinct ISO codes, but rather as regional groupings of multiple varieties~\cite{LOdP2018}. Linguistically, there are two major groups of Quechuas: Quechua I and Quechua II, and several subgroups~\cite{HuamanLCH23,Torero64}, and most of these Quechua-speaking populations are located within Peruvian territory. According to the 2017 National Census in Peru~\cite{LOdP2018}, the country has approximately 3805531 Quechua speakers, and in the Puno region alone there are 474203 Quechua speakers.

This paper aims to present an overview of Quechua language speech datasets within the Mozilla Common Voice platform\footnote{\url{https://commonvoice.mozilla.org/}} and to detail the integration of the Puno Quechua (qxp) into the platform. Furthermore, we explore its current statistical landscape, the Puno Quechua integration, and discuss a research agenda related to ethical considerations and community engagement.

The remainder of the paper is organized as follows. Section~\ref{sec:quechua-languages-in-cv} provides an overview of Quechua languages in the Common Voice platform. Section~\ref{sec:puno-quechua} and~\ref{sec:results} describe the process and results of integrating Puno Quechua. Finally, Section~\ref{sec:conclusions} concludes the paper with some remarks and future work plans.

%% file: chapter/2-quechua-languages.tex
\section{Quechua Languages in Common Voice}
\label{sec:quechua-languages-in-cv}
This section discusses the integration of Quechua languages into Common Voice, highlighting the platform's collaborative framework, Quechua linguistic diversity, and current status of contributions (see Figure~\ref{fig:languages}).

\subsection{Common Voice: A Collaborative Framework for Speech Data}
\label{subsec:common-voice}
The Common Voice project is an initiative to make speech technology accessible to everyone, especially to under-resourced languages. It started in 2017 with an emphasis on the English language, since 2018 was made available to any language. The operational framework is entirely reliant on volunteer contributions: users collect and upload sentences, then sentences are recorded, afterwards sentences are validated by others. This process allows anyone to contribute with more sentences, recordings, and validations~\cite{ArdilaBDKMHMSTW20}, ensuring a quality control over the data. The resulting dataset consists of MP3 files with their corresponding texts, if available, including demographic metadata like age, sex, and accent.

\begin{figure}[h]
    \centering
    \includegraphics[width=0.9\linewidth]{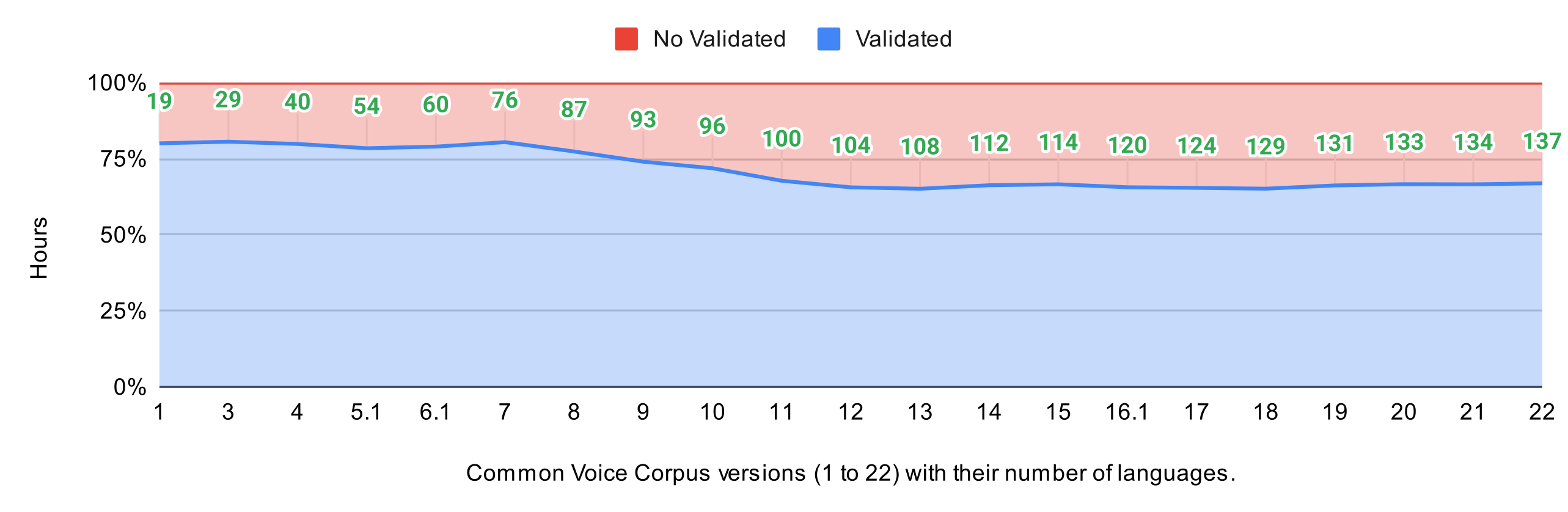}
    \caption{Language coverage by Common Voice versions.}
    \label{fig:languages}
\end{figure}

\begin{figure}
    \centering
    \includegraphics[width=0.98\linewidth]{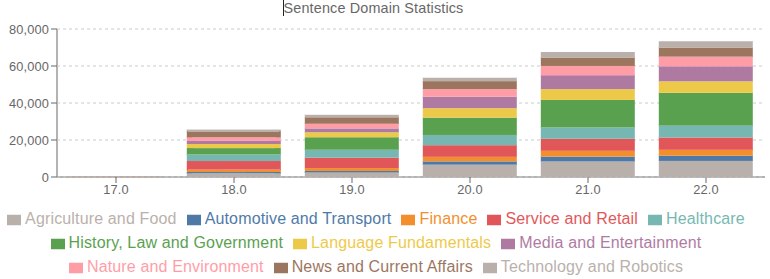}    
    \caption{Sentence domain distribution in Common Voice versions.}
    \label{fig:sentences-domain}
\end{figure}

Since its inception, Common Voice has grown significantly. The Common Voice corpus expanded from 1368 hours across 19 languages in version 1 to 33815 hours across 137 languages in version 22\footnote{\url{https://github.com/common-voice/cv-dataset/blob/main/README.md}}, see Figure~\ref{fig:languages}. Starting from version 17, sentences are categorized into 11 thematic domains, see Figure~\ref{fig:sentences-domain}. All collected data is released under a Creative Commons CC0 license, which fosters development and innovation in voice-enabled technologies.

\subsubsection{Integrating a language into Common Voice.}
\label{subsec:lang-in-cv}
It involves two main steps: 
\begin{enumerate}
    \item \textbf{Language Onboarding}: A language is proposed via GitHub by opening an issue with language-specific details\footnote{\url{https://github.com/common-voice/common-voice}}. Once it is approved, a localization project is created on Pontoon\footnote{\url{https://pontoon.mozilla.org/}} to translate the Common Voice platform. 
    \item \textbf{Corpus Collection}: A set of sentences with CC0 licesing is required for people to read aloud. After this is uploaded and approved, the language is officially launched on the Common Voice platform.
\end{enumerate}
Once integrated, community members can begin recording and validating voice clips. Language data is typically released as part of the Common Voice corpus within three months\footnote{\url{https://commonvoice.mozilla.org/en/about}}.

\subsection{Quechua Languages}
\label{subsec:quechua-languages}
Quechua languages, 43 identified on Ethnologue\footnote{\url{https://www.ethnologue.com/language/que/}} and 44 identified on Glottolog\footnote{\url{https://glottolog.org/resource/languoid/id/quec1387}}, are deeply rooted in South America. Quechua speakers are distributed across Argentina, Bolivia, Chile, Colombia, Ecuador, and Peru, with an estimated 7 million native speakers overall, and in Peru alone 3805531 quechua speakers~\cite{LOdP2018}.
Currently, 17 Quechua languages are supported on Common Voice, each identified by an ISO 639-3 Code\footnote{\url{https://iso639-3.sil.org/code/que}}. These languages are at various stages of integration: language onboarding or corpus collection, see Section~\ref{subsec:lang-in-cv}.

\begin{figure}[h]
    \centering
    \includegraphics[width=0.98\linewidth]{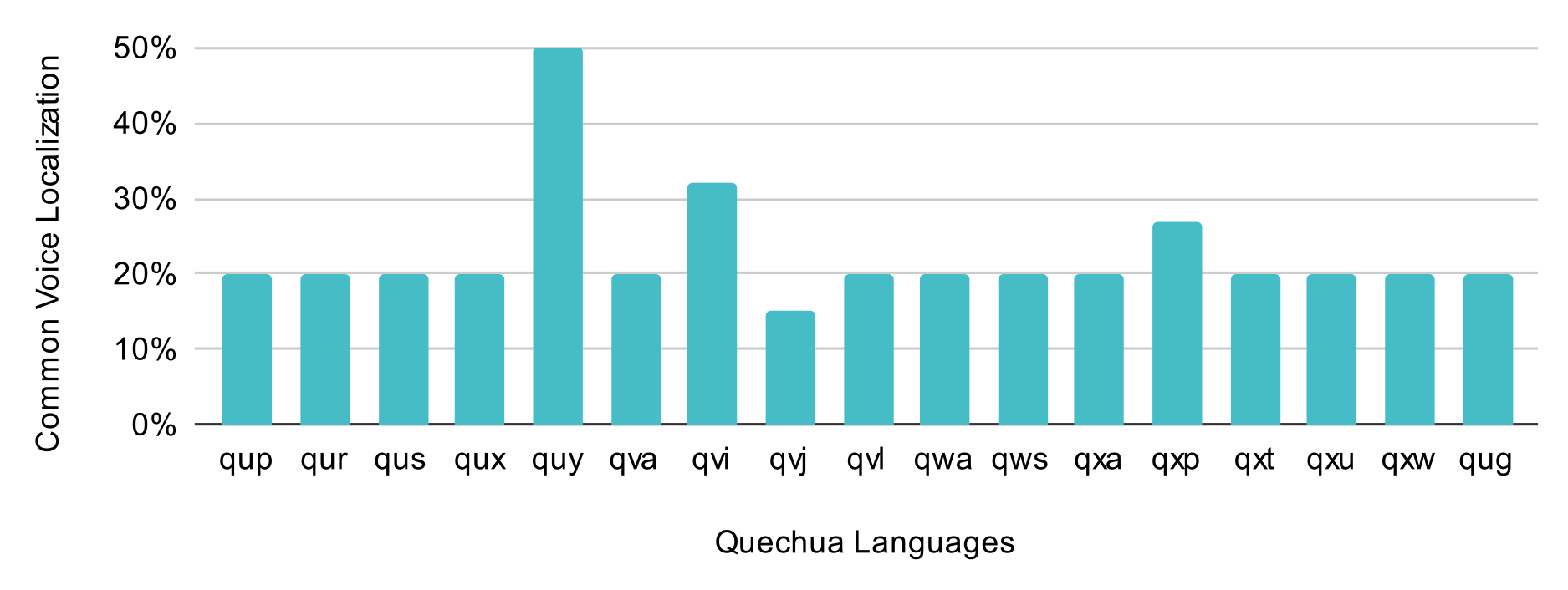}
    \caption{Localization efforts by Quechua communities in Common Voice platform.}
    \label{fig:quechua-languages}
\end{figure}

Figure~\ref{fig:quechua-languages} shows localization efforts by Quechua-speaking communities to integrate their language into Common Voice. Most languages have localized the minimum 20\% localization threshold required for basic functionality and reading-based speech recording. However, 75\% is the minimum required to make the user-platform interaction more correct and to support spontaneous speech recording. 

\begin{figure}
    \centering
    \includegraphics[width=0.98\linewidth]{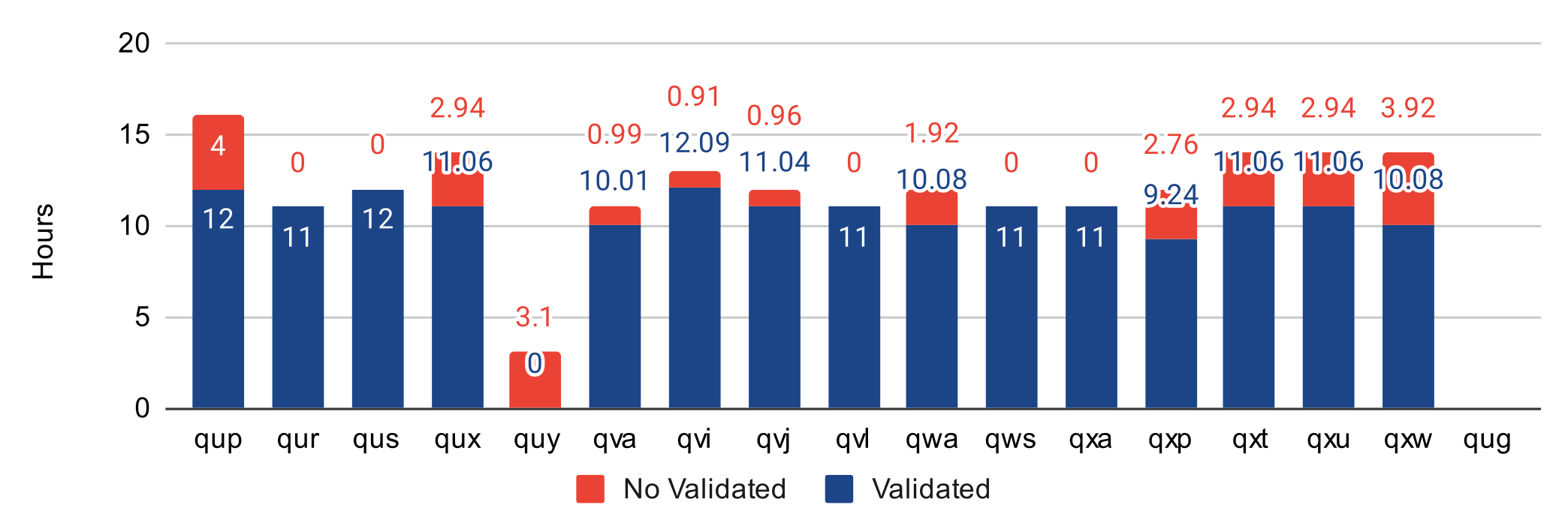}
    \caption{Speech hours by Quechua language in Common Voice.}
    \label{fig:quechua-hours}
\end{figure}

The quantity of recorded speech varies widely among Quechua languages. For example, qug (Chimborazo Highland Quichua) has 0 hours recorded, while qup (Southern Pastaza Quechua) has 16 hours, see Figure~\ref{fig:quechua-hours}. Overall, approximately 86\% of Quechua submitted recordings have been validated.

%% file: chapter/3-case-study.tex
\section{Case Study: Puno Quechua}
\label{sec:puno-quechua}
This section presents a case study on the integration of Puno Quechua into the Common Voice platform. It begins by characterizing Puno Quechua linguistically and demographically, followed by a detailed account of the steps taken to integrate it into Common Voice, including platform localization, sentence collection, and speech recording efforts.

\subsection{Puno Quechua Language}

Puno Quechua, identified by the ISO 639-3: qxp\footnote{\url{https://iso639-3.sil.org/code/qxp}}, belongs to the Quechua II group. According to the Peruvian government, it is further sub-grouped under Southern Quechua, specifically within the Quechua Collao. The 2017 National Census~\cite{LOdP2018} estimates approximately 474203 speakers of Puno Quechua, predominantly in the Puno region of Peru. Historically, the number of Quechua speakers in this region has shown a significant decline: from 87.23\% in 1940 to 50\% in 1993, and further to 38.01\% in 2007~\cite{Andrade2019,CarbajalGHMRCV2019,Lopes1993}.

According to Glottolog, most linguistic resources for Puno Quechua were published between 1963 and 1993, with few notable exceptions from 2005. These resources offer valuable insights into the language's linguistic characteristics. Linguistically, Puno Quechua adheres to the agglutinative structure characteristic of all Quechua languages, where words are formed by adding multiple suffixes to a root word, and follows a Subject-Object-Verb (SOV) sentence structure.

\begin{figure}[h]
    \centering
    \includegraphics[width=0.5\linewidth]{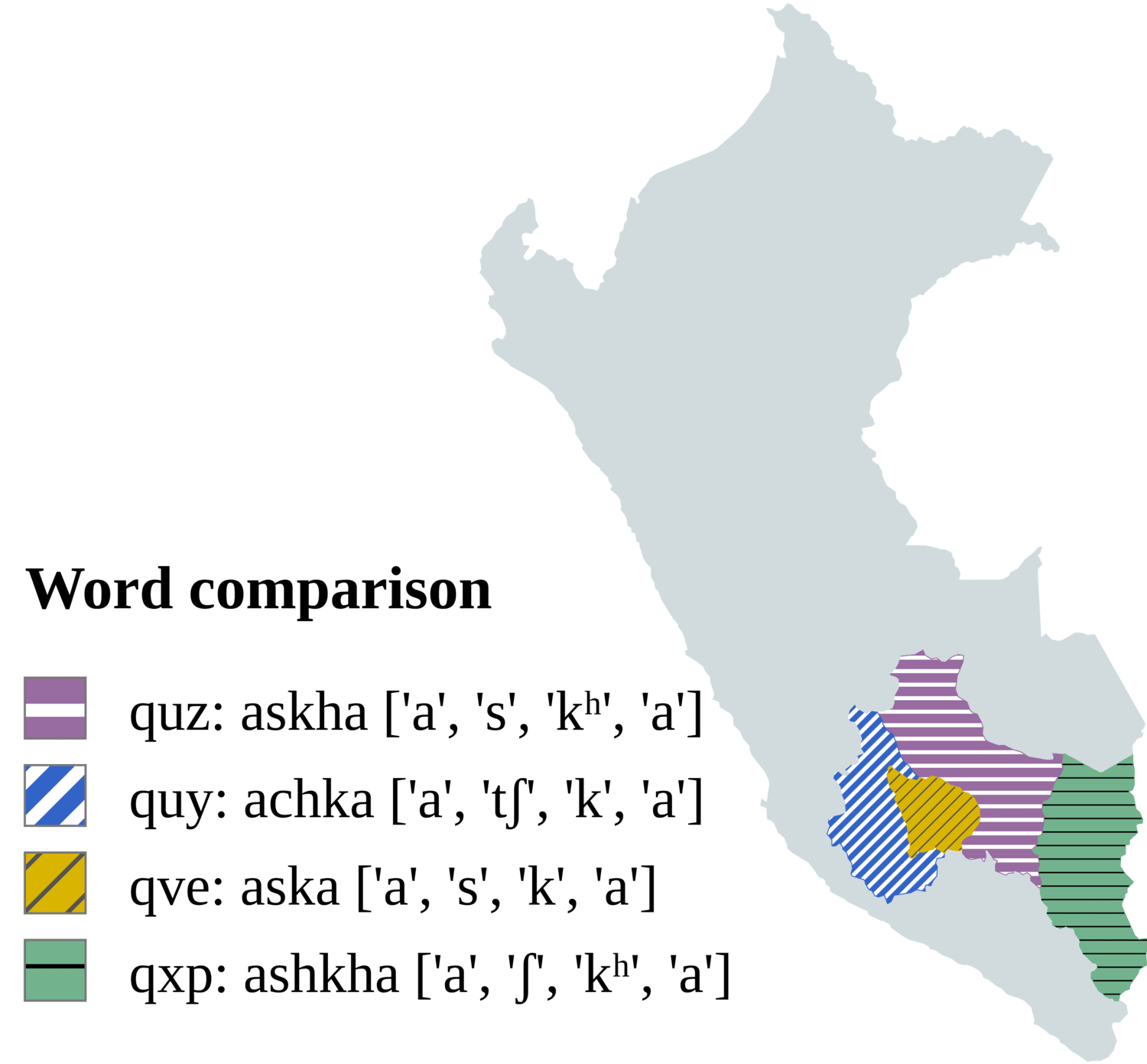}
    \caption{Comparison of a word across Southern Quechua languages.}
    \label{fig:quechua-word}
\end{figure}

A comparative lexical analysis across different Southern Quechua varieties, as illustrated in Figure~\ref{fig:quechua-word}, demonstrates lexical variation\footnote{Cross-Linguistic Data Formats dataset \url{https://cldf.clld.org/}}. For instance, the English word \textit{a lot} is represented as \textit{ashkha} in Puno Quechua (qxp), but varies in other Southern Quechuas. This lexical diversity underscores the need for separate speech datasets for each Quechua language, ensuring linguistic accuracy and utility for speech technology development.

\subsection{Integrating Puno Quechua into Common Voice}
The integration of Puno Quechua into Common Voice involved two tasks: language onboarding and corpus collection.
\subsubsection{Language Onboarding:}
This task entailed requesting our language via GitHub, where we opened an issue with Puno Quechua language information. Once it was approved, we got access to a Common Voice Pontoon project, where we started translating and localizing the interface into Puno Quechua. Our efforts resulted in a 27\% translation rate, which indicates that 466 out of 1793 total strings have been translated\footnote{\url{https://pontoon.mozilla.org/qxp/common-voice/}}. This partial localization covers both Reading Speech and Spontaneous Speech interfaces.
\subsubsection{Corpus Collection:}
\begin{figure}
    \centering
    \includegraphics[width=0.9\linewidth]{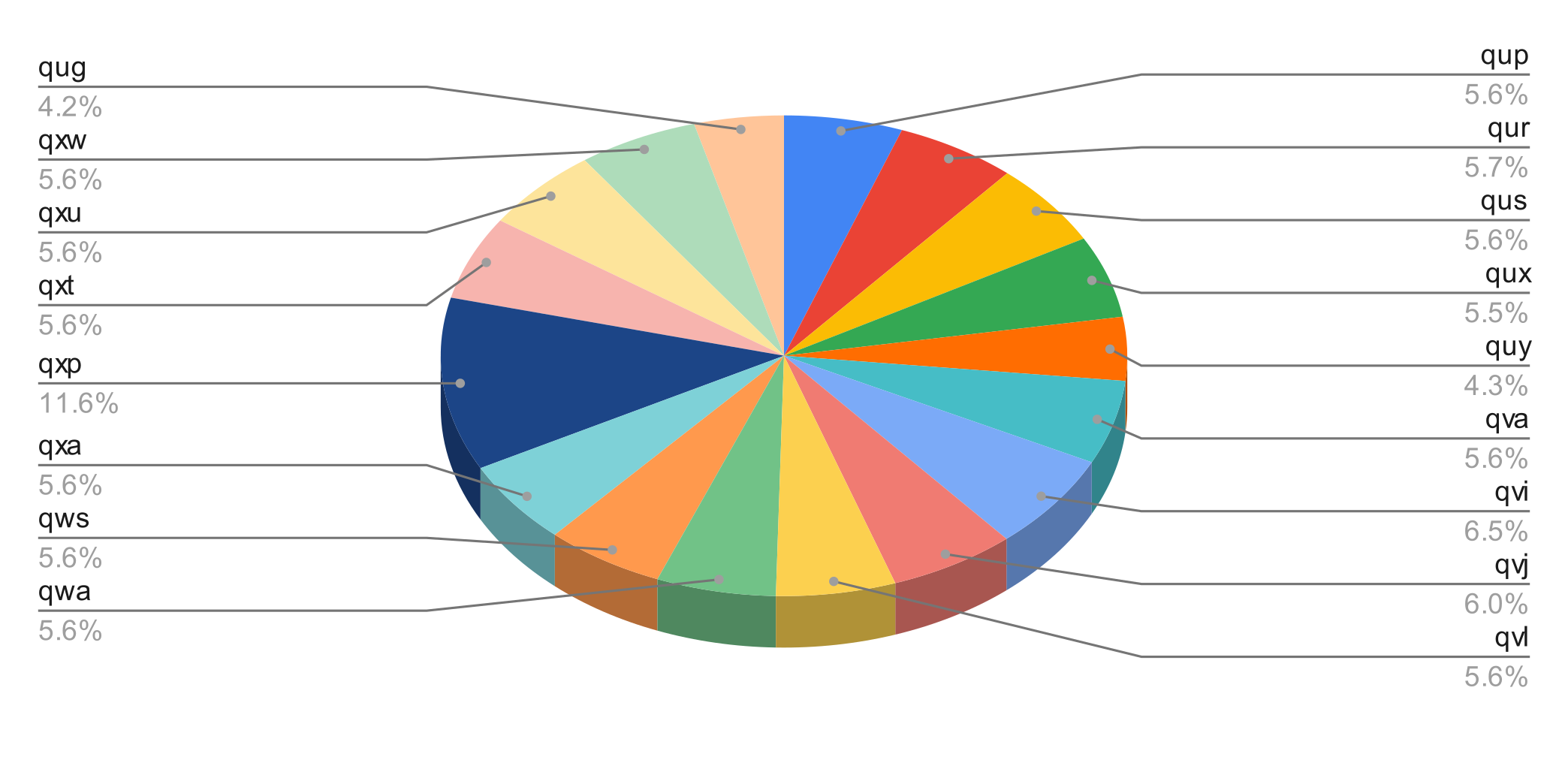}
    \caption{Quechua sentences coverage by Quechua languages.}
    \label{fig:quechua-sentences}
\end{figure}

This phase focused on gathering CC0-licensed sentences suitable for the platform. We collected 2065 sentences and subsequently uploaded them to the Common Voice platform. These Puno Quechua sentences represent 11.6\% of the total Quechua sentences available on the platform (see Figure~\ref{fig:quechua-sentences}).

\subsection{Speech Recording}
\label{subsec:reading-speech}
Speech recording efforts for Puno Quechua within Common Voice were divided into Reading Speech and Spontaneous Speech contributions.

\subsubsection{Reading Speech Recording:}
It involved volunteers reading aloud the uploaded sentences and recording them. We try our best to contribute with clear pronunciation and in noise-free environment, so the recordings are of high quality. So far, community validation has reached 77\% (9 hours and 24 minutes) of the total recorded clips (12 hours), see Figure~\ref{fig:quechua-hours}. We highlight the importance of sustained community engagement to improve the quality and robustness of the Puno Quechua speech dataset.

\subsubsection{Spontaneous Speech Recording:}
This more advanced contribution type involves community members answering open-ended questions, which are then transcribed and validated. We uploaded 40 questions specifically within the Agricultural and Food domain for the community to answer. The transcription process for spontaneous speech also accounts for casual speaking patterns, including interjections like ``mm'' and ``ahh'', as well as repetitions. Subsequently, these transcriptions are validated against the speech recordings by other speakers. This approach captures diverse speaking styles, including accents, hesitations, and informal phrasing, which are vital for developing robust and adaptable voice technologies.

%% file: chapter/4-results.tex
\section{Results}
\label{sec:results}
This chapter presents the key outcomes of integrating Quechua languages, with a specific focus on Puno Quechua, into the Common Voice platform. It highlights the current status of the collected speech corpus and outlines a comprehensive research agenda and social framework necessary for advancing the development of high-quality speech datasets for under-resourced languages.

\subsection{Current Status of Quechua Languages in Common Voice}

Following the integration of 17 Quechua languages into the Common Voice platform, particularly the Puno Quechua, this paper demonstrates a viable pathway for making under-resourced languages available as community-driven text-to-speech corpora. Currently, the Common Voice platform hosts 191.1 hours of speech data across all integrated Quechua languages, of which 163.72 hours (86\%) have been validated by the community. In the case of Puno Quechua, a total of 12 hours of speech were recorded, with 77\% of this data successfully validated. Table~\ref{tab:Quechua-on-CV} provides a summary of the current contributions.

\begin{table}[h!]
\centering
    \begin{tabular}{ |l|r|r|c|r|  }
        \hline
        &Hours&Speakers&Validation Progress&Sentences\\
        \hline
        Quechua Languages&191.1&238&86\%&17754\\
        \hline
        Puno Quechua&12.0&14&77\%&2065\\
        \hline
    \end{tabular}
\caption{Overview of Quechua Languages Data on Common Voice.}
    \label{tab:Quechua-on-CV}
\end{table}

\subsection{Research Agenda for High-Quality Quechua Speech Datasets}
To ensure the creation of high-quality Quechua speech datasets suitable for diverse applications such as speech recognition, text-to-speech (TTS) synthesis, and language preservation, research challenges and opportunities must be addressed. This research agenda outlines, integrating the technical requirements of the Common Voice platform, the linguistic complexities of under-resourced languages, and the socio-cultural dynamics required for community engagement.
\subsubsection{Challenge: Language Support and Orthographic Standards.}
It involves ensuring that all languages, regardless of their listing on Glottolog or ISO, are adequately supported. This requires review to ensure localizations and collected sentences adhere to established orthographic standards.
    \textbf{Opportunity}: Collaborating with universities, research institutes, and Quechua-speaking organizations can facilitate the quality integration of any language into Common Voice by leveraging their linguistic expertise and community ties.
\subsubsection{Challenge: Building Diverse and Culturally Relevant Text Corpora.}
This includes ensuring a balanced distribution of sentences across various domains to mitigate bias in the resulting speech dataset.
    \textbf{Opportunity}: Actively mobilizing the community in the sentence collection process is crucial for ensuring that gathered sentences are culturally appropriate and domain-relevant, thereby reducing inherent biases.
\subsubsection{Challenge: High-Quality Voice Contributions.}
It involves effective training and continuos support for community contributors. Practical barriers identified during Puno Quechua data collection include expensive internet access and limited digital literacy, which hinder community participation. 
    \textbf{Opportunity}: A hybrid approach combining online Common Voice contributions with organized offline recordings campaigns is essential. This strategy should include the provision of devices, internet access, offline access to the platform, alongside localized training workshops, to maximize participation and overcome socio-economic barriers.
\subsubsection{Challenge: Spontaneous Speech and Code-Switching.}
It involves capturing spontaneous speech ``on the fly'', which brings unique challenges due to linguistic phenomena that deviate from carefully read text, such as prolonged pauses, false starts, repetitions, and background noise. An additional complexity arises from the frequent code-switching between Quechua and Spanish among bilingual speakers.
    \textbf{Opportunity}: The development of offline mobile applications can democratize data collection, enabling participation from a wider range of speakers and in diverse contexts. This directly addresses the need for diverse datasets to reduce bias in speech recognition models. For Puno Quechua, we focus our effort to minimize code-switching by collecting speech centered around culturally vibrant topics such as agriculture, farming, and daily life.

\subsection{Social framework and ethical considerations}
Beyond the technical research agenda, the sustainable integration of under-resourced languages into platforms like Common Voice necessitates a robust social framework grounded in ethical considerations:
\begin{itemize}
    \item \textbf{Prioritize Localization and Community Building:} Initial localization efforts for activating a language on the platform. Simultaneously, it is necessary to build relationships with Quechua communities, linguists, research institutions, and public organizations to foster sustainable engagement.
    \item \textbf{Ethical Framework and Data Sovereignty:} It is essential to establish a solid ethical framework that goes beyond mere regulatory compliance. This includes implementing informed consent processes that are culturally sensitive. Furthermore, it is crucial to engage with language communities to ensure a comprehensive understanding of the Creative Common CC0 licensing towards their data sovereignty. This engagement should explore the potential for community-centric data licenses or benefit sharing agreements that align with community values.
    \item \textbf{Importance of Spontaneous Speech:} The inclusion of spontaneous speech is crucial for training real-world speech recognition applications, as most voice assistants require the ability to understand natural and unscripted speech to be effective in daily use.
\end{itemize}

By addressing the outlined research agenda and ethical considerations, the integration of languages into Common Voice can establish a robust foundation, empowering language communities to contribute significantly. This paper brings insights into the digital preservation and revitalization of under-resourced languages through their integration into the Common Voice ecosystem.

%% file: chapter/5-conclusion.tex
\section{Conclusions}
\label{sec:conclusions}
This paper demonstrates the integration of under-resourced Quechua languages into Mozilla's Common Voice platform, showcasing a viable model for creating vital speech datasets. We highlighted the contribution of 191.1 hours of Quechua speech (86\% validated), including 12 hours for Puno Quechua (77\% validated), underscoring Common Voice's potential to address speech data scarcity.

To advance these efforts, a comprehensive research agenda is crucial, focusing on ensuring linguistic accuracy through distinct datasets, developing diverse text corpora, and overcoming digital access barriers via hybrid online-offline approaches for high-quality, spontaneous speech collection. Furthermore, a robust social framework is essential, emphasizing community building, culturally sensitive ethical practices, and discussions on indigenous data sovereignty.

Our future work will be focused on expanding the Common Voice dataset to include all remaining Quechua languages, while ensuring quality, diversity, and sustainability throughout community contributions. Additionally, we will explore ways to enable offline contributions through integrations with instant messaging applications, allowing people to contribute from
different devices and connection modes.

%% file: chapter/6-acknowledgment.tex
\begin{credits}
\subsubsection{\ackname} 
We thank the Mozilla Foundation, the Common Voice project, and the Mozilla Common Voice community for their support. A deep gratitude to Francis Tyers, for his unconditional support in overcoming challenges during the Speech recording campaigns. Our sincere appreciation to the T'ikariy community in Nuñoa, Puno, Peru, specially Christian Tapara, Eliana Quispe, Luis Huaman, Magaly Mullisaca, Maribel Hancco, Milagros Huaman, Nely Quispe, and everybody who contributed with their voices to this project.
\end{credits}

%% file: 42.bbl
\begin{thebibliography}{10}
\providecommand{\url}[1]{\texttt{#1}}
\providecommand{\urlprefix}{URL }
\providecommand{\doi}[1]{https://doi.org/#1}

\bibitem{Andrade2019}
Andrade~Ciudad, L.: Ten news about quechua in the last peruvian census. Letras (Lima)  \textbf{90},  41--70 (2019), \url{http://dx.doi.org/10.30920/letras.90.132.2}

\bibitem{ArdilaBDKMHMSTW20}
Ardila, R., Branson, M., Davis, K., Kohler, M., Meyer, J., Henretty, M., Morais, R., Saunders, L., Tyers, F.M., Weber, G.: Common voice: {A} massively-multilingual speech corpus. In: Proceedings of The 12th Language Resources and Evaluation Conference, {LREC} 2020, Marseille, France, May 11-16, 2020. pp. 4218--4222. European Language Resources Association (2020), \url{https://aclanthology.org/2020.lrec-1.520/}

\bibitem{BesacierBKS14a}
Besacier, L., Barnard, E., Karpov, A., Schultz, T.: Automatic speech recognition for under-resourced languages: {A} survey. Speech Commun.  \textbf{56},  85--100 (2014), \url{https://doi.org/10.1016/j.specom.2013.07.008}

\bibitem{CarbajalGHMRCV2019}
Carbajal~Solis, V., Garcia~Rivera, F.A., Huamancayo~Curi, E.Y., Mori~Clement, M., Rodriguez~Aguero, M., Gutierrez, C., Verastegui~Walqui, N.: Lenguas Originarias del Peru. Ministerio de Educacion del Peru (2019), \url{https://hdl.handle.net/20.500.12799/6261}

\bibitem{HuamanHH22}
Huaman, E., Huaman, J.L., Huaman, W.: Getting quechua closer to final users through knowledge graphs. In: Information Management and Big Data - 9th Annual International Conference, SIMBig 2022, Lima, Peru, November 16-18, 2022, Proceedings. Communications in Computer and Information Science, vol.~1837, pp. 61--69. Springer (2022), \url{https://doi.org/10.1007/978-3-031-35445-8\_5}

\bibitem{HuamanHH23}
Huaman, E., Huaman, W., Huaman, J.L.: Making an under-resourced language available on the wikidata knowledge graph: Quechua language. In: Information Management and Big Data - 9th Annual International Conference, SIMBig 2024, Ilo, Peru, November 20-22, 2024, Proceedings. Communications in Computer and Information Science, vol.~2496, pp. 212--224. Springer (2024). \doi{10.1007/978-3-031-91428-7\_15}, \url{https://doi.org/10.1007/978-3-031-91428-7\_15}

\bibitem{HuamanLCH23}
Huaman, E., Lindemann, D., Caruso, V., Huaman, J.L.: {QICHWABASE:} {A} quechua language and knowledge base for quechua communities. CoRR  \textbf{abs/2305.06173} (2023). \doi{10.48550/ARXIV.2305.06173}

\bibitem{Lopes1993}
L{\'o}pes, L.E.: Educaci{\'o}n biling{\"u}e en puno-per{\'u}: activos y pasivos de un programa de educati{\'o}n rural en los andes. Ling{\"u}{\'\i}stica ind{\'\i}gena e educa{\c{c}}{\~a}o na Am{\'e}rica Latina. Campinas: UNICAMP pp. 13--70 (1993), \url{{https://etnolinguistica.wdfiles.com/local--files/biblio%3Alopez-1993-educacion/Lopez_1993_EducacionBilinguePunoPeru.pdf}}

\bibitem{LOdP2018}
{Ministerio de Educación del Perú}: {Lenguas originarias del Perú}. {Biblioteca Nactional del Perú}, Lima, Per{ú} (2018)

\bibitem{Torero64}
Torero, A.: Los dialectos quechuas. In: Anales Cientificos, Volume II, Lima, 1964. pp. 446--478. Universidad Agraria Peru (1964)

\end{thebibliography}
